\documentclass[runningheads]{llncs}
\usepackage[T1]{fontenc}
\usepackage{graphicx}
\usepackage{url}
\usepackage{subfig}
\usepackage{cite}
\usepackage{amsmath,amssymb,amsfonts}
\usepackage{algorithmic}
\usepackage{eucal}
\usepackage{textcomp}
\usepackage{adjustbox}
\usepackage{xcolor}
\usepackage[textsize=tiny,textwidth=1.4cm]{todonotes}
\usepackage{mathtools,lipsum, nccmath,cuted}
\usepackage{siunitx}

\usepackage{comment}
\usepackage{makecell}
\usepackage{algorithm}
\usepackage{algorithmic}
\usepackage{gensymb}
\usepackage{ulem}
\usepackage{colortbl}
\usepackage{hyperref}
\usepackage{enumitem}
\usepackage{booktabs}

\begin{document}
\title{Fusion of ML with numerical simulation for optimized propeller design
\thanks{preprint submitted to AIAI 2023.}}
%
%
%
\author{Harsh vardhan \and
Peter Volgyesi \and
Janos Sztipanovits}

\authorrunning{Vardhan et al.}
%
\institute{Insititue of Software and Integrated System, Vanderbilt University 
\email{harsh.vardhan@vanderbilt.edu, peter.volgyesi@vanderbilt.edu, Janos.sztipanovits@vanderbilt.edu}}
\maketitle              

\begin{abstract}
\textit{In computer-aided engineering design, the goal of a designer is to find an optimal design on a given requirement using the numerical simulator in loop with an optimization method. In this design optimization process, a good design optimization process is one that can reduce the time from inception to design. In this work, we take a class of design problem, that is computationally cheap to evaluate but has high dimensional design space. In such cases, traditional surrogate-based optimization does not offer any benefits. In this work, we propose an alternative way to use ML model to surrogate the design process that formulates the search problem as an inverse problem and can save time by finding the optimal design or at least a good initial seed design for optimization. By using this trained surrogate model with the traditional optimization method, we can get the best of both worlds.
We call this as \textit{Surrogate Assisted Optimization (SAO)}- a hybrid approach by mixing ML surrogate with the traditional optimization method. Empirical evaluations of propeller design problems show that a better efficient design can be found in fewer evaluations using SAO. 
}
\end{abstract}

\section{Introduction}
\label{sec:intro}
In the last decades, considerable effort has been made to rapidly optimize the designs in different engineering problems \cite{martins2021engineering}\cite{vardhan2021machine}\cite{vardhan2022deep}. The main bottleneck in rapid design optimization is either slow evaluation due to a complex numerical simulation process or high-dimensional design space or both. This high dimensional design space can be due to ranges of search of independent variables or a large number of such variables or both. In case of problems that involve complex numerical models and simulation processes, Surrogate-based optimization (SBO)\cite{sobester2008engineering} is the main approach, where a data-driven learning model is trained to replace numerical simulation in the optimization loop\cite{sobester2008engineering}\cite{vardhan2022deepal}. The motivation for creating a surrogate is cheap approximate evaluation in comparison to direct numerical simulation. There are other cases where due to the availability of coarse approximate physics models, numerical simulations are cheap to evaluate and the only challenge arises from high dimensional design space. The traditional SBO does not offer much in these cases. In this work, we try to address this class of problems by exploring the possibility of using ML in these problems and its benefits during the design optimization process. For this purpose, we propose a surrogate-assisted optimization (SAO), where the surrogate is trained on earlier collected labeled data from the design and requirement space. By capitalizing on the generalization capability of trained ML models, we want to speed up the design process for a range of requirements. In such cases, the trained surrogate acts as a memory of experience (similar to an expert human designer) and is used to find good design directly or at least provide a good seed design for further optimization. For this purpose, the surrogate uses both nonlinear interpolation and nonlinear mapping to provide a good baseline for further optimization. The challenge of creating a surrogate in this case arises due to modeling expectations in this case. The modeling expectation is to try to get a good design from the requirement directly. Due to the acausal relationship between the requirement on design and the design parameter, it must be modeled as an inverse problem. Due to the causality principle, the forward problem in engineering systems has a distinct solution. On the other hand, the inverse problem might have numerous solutions if various system parameters predict the same effect.  Generally, the Inverse modeling problem is formalized in a probabilistic framework which is complex and not very accurate for high dimensional input-output and design space. We attempt this problem from geometric data summarizing algorithms that can model inverse problems and are useful in these problems. 
To differentiate this approach from surrogate-based optimization (SBO), we call this approach \textbf{Surrogate-Assisted-Optimization (SAO)}. The main difference between SBO and SAO is that in SBO, we use a surrogate in the optimization loop while in SAO, a surrogate is external to the optimization loop and only used to get a good initial baseline, further design optimization starts with this initial seed design provided by surrogate. The other difference is, in SAO surrogate attempts to inverse modeling problem instead of forward modeling problem in SBO. In SAO, the role of a surrogate is to provide all possible good designs or seed designs.    
For surrogate modeling, our choice of models are random forest and decision tree. The random forest has empirically shown to work the best for inverse modeling problems \cite{aller2023study}. We also selected to train one decision tree on the entire data to create a memory map of collected data. Empirically we observed adding one decision tree trained on the entire data set along with a random forest of decision trees trained on various sub-samples of the dataset and using averaging improves the predictive accuracy and control over-fitting. 

For empirical evaluation, we take the use case problem of propeller design\cite{vardhan2021machine}, and the design space after coarse discretization is of the order of approximately  $10^{38}$. Based on the collected data requirement and  training, when the SAO approach is applied to multiple optimization problems sampled from the requirement space. In all cases, we found SAO that leverage on initial good seed design from surrogate can find a better design on a given budget in comparison to the traditional method.  

\section{Background and Problem Formulation}
\label{sec:background}

\subsection{Background}
\subsubsection{Propeller:}
\begin{figure}[tbhp]
\centering
   \includegraphics[width=1.0
   \linewidth]{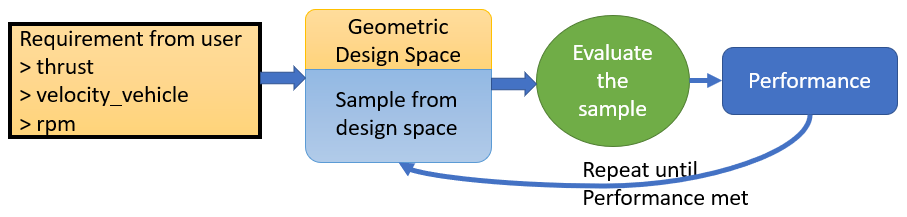}
   \caption{Propeller design optimization process in openProp. Sample evaluation is done in openProp simulator and performance is measured by the efficiency of the propeller.}
   \label{fig:problem}
\end{figure}

Propellers are mechanical devices that convert rotational energy to thrust by forcing incoming forward fluids axially toward the outgoing direction. On a given operating condition such as the advance ratio ($J$) rpm of the motor and desired thrust, the performance of a propeller is characterized by its physical parameters such as the number of blades ($Z$), diameter of the propeller ($D$), chord radial distribution ($C/D$), pitch radial distribution ($P/R$) and hub diameter($D_{hub}$)\cite{epps2009openprop, vardhan2021machine}. The goal of a propeller designer is to find the optimal geometric parameters that can meet this thrust requirement with maximum power efficiency ($\eta$) (refer Figure~\ref{fig:problem}). 
We use openprop\cite{epps2009openprop} as our numerical simulation tool in this work. The output of simulation informs about the quality of the design choice, and accordingly, a bad design choice may result in poor efficiency or infeasible design and vice versa. The biggest challenge in the design search process arises from the exponentially large design space of the geometric parameter. 

\subsubsection{Openprop:}
Openprop is a propeller design tool based on the theory of the moderately loaded lifting line, with trailing vorticity oriented to the regional flow rate. Optimization processes in openprop involve solving the Lagrange multiplier ($\lambda_1$) for finding the ideal circulation distribution along the blade's span given the inflow conditions and blade $2D$ section parameters. The openprop applies Coney's formulation\cite{coney1989method} to determine produced torque $Q$, thrust $T$, and circulation distribution $Gamma$ for a given required thrust $T S$.
For optimization purposes, an auxiliary function is defined as follows:
\begin{equation}
    H= Q+ \lambda_1 (T- T_s)
\end{equation}
If $T=T_S$ then a minimum value of $H$ coincides with a minimum value of $Q$. To find the minimum, the partial derivative with respect to unknowns is set to zero.
\begin{gather}
    \frac{\partial H}{\partial\Gamma(i)}=0\; for\; i=1,2,...,M \label{eq:prop1} \\
    \frac{\partial H}{\partial \lambda_1}=0 \label{eq:prop2}
\end{gather}
By solving these $ M$ systems of non-linear equations using the iterative method -i.e. by thawing other variables and linearizing the equations with unknowns {$\hat{\Gamma}, \hat{\lambda_1}$}, an optimal circulation distribution and a physically realistic design can be found. For more details on numerical methods, refer to \cite{epps2009openprop, coney1989method}. 
\begin{figure}[tbhp]
\centering
   \includegraphics[width=1.0
   \linewidth]{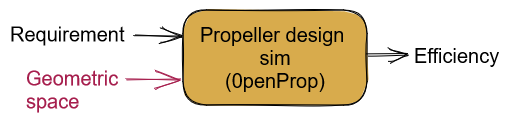}
   \caption{OpenProp Numerical Simulation}
   \label{fig:Approach}
\end{figure}

\subsubsection{Random forest and Decision tree:}
A random forest\cite{breiman2001random} is a non-parametric supervised machine learning method that is an ensemble of various decision trees. Each decision tree is a machine-learning model that can be trained for regression and classification purposes. The fundamental of random forest learning is bagging \cite{breiman2017classification, vardhan2022reduced}, in which the decision tree algorithm is applied multiple times on a subset of data and then the output result is averaged. 
The goal is to train many uncorrelated trees by sub-sampling $D$ data points with replacement from a data-set $X$.  This process reduces over-fitting by averaging the prediction from different trained on different data sets sampled from the same data distribution. A decision tree is created by recursive binary partitioning of the variable space until the partitioning of the space is complete. 
\vspace{-2mm}
\subsection{Problem Formulation}
Based on a given requirement imposed on a design in terms of operational and performance conditions, the goal of a designer is to find an optimal geometric parameter of the propeller in minimum time.  In OpenProp,  the input design space can be split into two parts: (1) \textbf{Requirement space ($\mathcal{R}$)} that comprises of thrust, velocity of vehicle, rpm, and (2) \textbf{Geometric design space ($\mathcal{G}$)} comprises of chord profile radial distribution ($C/D$), diameter ($D$), hub diameter ($Dhub$), etc). The design space considered for this study is taken from \cite{vardhan2021machine}. Once samples were taken from this space, the requirement, and geometric design are put in the iterative numerical simulation algorithm to find the efficiency ($\eta$) of the design. The goal of design optimization is formalized as : 
\begin{gather}
\underset{g \in \mathcal{G}}{\mathrm{argmax}} \;\eta \;\; for\; a\; given\; r \sim \mathcal{R} 
\end{gather}   
Since this design optimization process for a given requirement involves running a sequential design selection from the input geometric space ($\mathcal{G}$), its evaluation and optimization until the requirements are satisfied. In such a case, another important aspect is to reduce the inception to design time ($\mathcal{T}_{design}$) i.e. design optimization time. Collectively, it can be written as: 
\begin{gather}
\underset{g \in \mathcal{G}}{\mathrm{argmax}} \;\eta \;\; for\; a\; given\; r \sim \mathcal{R} \\
min \;\mathcal{T}_{design}
\end{gather}


\section{Approach}
\label{sec:approach}
\subsection{Formulating design search as inverse problem:}
In forward modeling and prediction problems, we use a physical theory or simulation model for predicting the outcome ($\eta$) of parameter ($g$) defining a design behavior. The optimization process in the forward problem involves sampling from parameter space ($\mathcal{G}$) and striving to find the best parameter ($g^*$) that meets the requirement on the performance metrics ($\eta$). 
In the reciprocal situation, in inverse modeling and prediction problem, the values of the parameters representing a system  are inferred from values of the desired output and the goal is to find the desired values of the parameters ($g^*$) that represent the output ($\eta$) directly. 

In the propeller design use case,  the objective of a designer is to determine the best geometric characteristics of the propeller in minimum time, based on a particular demand imposed on the design in terms of operational and performance conditions.  
The inverse setting in this case has some unique features: 
\begin{enumerate}
    \item One part of the input variables is known i.e. requirement. The other part of the input is unknown (geometry).
    \item The effect or desired output is not fixed and the goal is to get the maximum possible efficiency that depends on requirements. ( for example, it is not possible to produce a thrust with a small rpm motor at some specific speed.)
\end{enumerate}
To address these situations we formulate our inverse modeling problem as selecting and training a prediction model that can map a given requirement to the geometry and efficiency.
  $$ \mathcal{IM}: \mathcal{R} \mapsto\{\mathcal{G},\eta\}   $$

Since it is not possible to find the maximum efficiency apriori, we filter all low-efficiency data sets (we treat these as infeasible designs) and keep only designs whose efficiency is higher. To model this inverse problem, we rely on geometric data summarizing techniques that learn the mapping between input and output space as sketches and the ability to regress between them. A sketch is a compressed mapping of output data set onto a data structure.

\subsection{Why random forest and decision tree is our choice for modeling this inverse problem?}
In the geometric data summarizing technique, the aim is to abstract data from the metric space to a compressed representation in a data structure that is quick to update with new information and supports queries. Let $D =\{d_1, d_2, ... , d_n\}$ are set of datapoints such that $d_i \in R^m $. For the purpose of representing data in sketches ($S$), the main requirement is the relationship ($\psi$) between the data points in metric space must be preserved in this data structure i.e  $\psi\{T(d_k,d_l)\}\approx \psi\{S(d_k,d_l)\} $.

One of the selected relationships ($\psi$) between datapoints in metric space is $L_p$ distance between datapoints.  In such case, a distance-preserving embedding of this relationship in metric space is equivalent to tree distance between two data points $d_k$ and $d_l$ in data structure ($S$). Tree distance is defined as the weight of the least common ancestor of $d_k$ and $d_l$\cite{guha2016robust}, then according to the Johnson-Lindenstrauss lemma\cite{lindenstrauss1984extensions} the tree distance can be bounded from at least $L_1 (d_k,d_l)$ to maximum $ O(d* log|k|/L_1(d_k,d_l))$. Accordingly, a point that is far from other points in the metric space will continue to be at least as far in a randomized decision tree. 

$$L_1 (d_k,d_l) \leq tree\;distance \leq O(d* log|k|/L_1(d_k,d_l))  $$
\textbf{Random Forest} is a collection of specific kind of decision tree where each tree in a random forest depends on the values of a random vector that was sampled randomly and with the same distribution for all the trees in the forest. When the number of trees in a forest increases, the generalization error converges to a limit. The strength of each individual tree in the forest and the correlation between them determine the accuracy of a forest of tree. The error rates are better than Adaboost when each node is split using a random selection of features\cite{breiman2001random}.
To create a tree ($h(x,\theta_k)$ in the forest, $\theta_k$ is independent identically distributed random vectors independent of the past random vectors $\theta_1,...,\theta_{k-1}$ but from the same distribution. Due to ensembling and randomness in the forest generation process, the variance in $tree\;distance$ also reduces to $L_1 (d_k,d_l)$. Accordingly, geometric summarization of data from metric space to random forest can maintains the $L_1$ norm between data points in expectation. The decision tree trained on entire data-set has over-fitting issue  and not suitable for generalization but due to space partitioning nature, it can map each observed requirement with multiple geometric designs and its efficiency when trained. By using both trained models in parallel, we can capitalize on both nonlinear mapping feature of decision tree as well as non linear regression/interpolation feature of random forest.   

\subsection{A hybrid optimization approach : Surrogate Assisted Optimization (SAO)}
\begin{figure*}[tbhp]
\centering
   \includegraphics[width=1.0
   \linewidth]{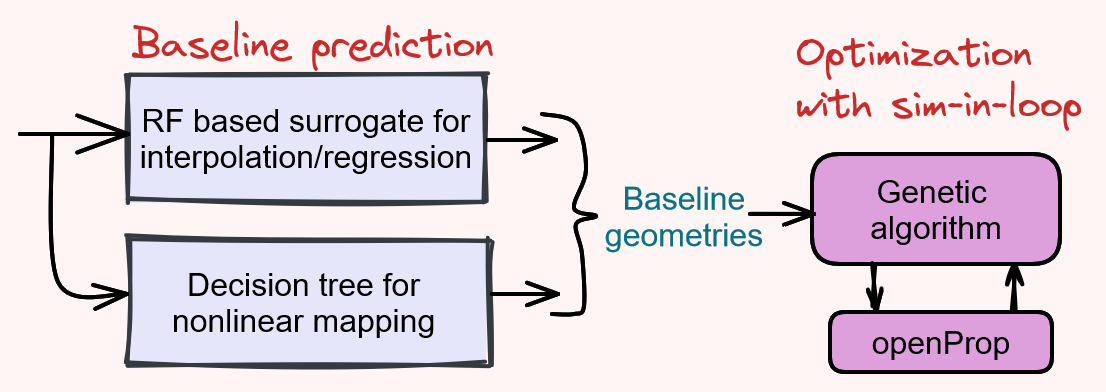}
   \caption{Surrogate Assisted Design optimization for propeller design}
   \label{fig:Approach}
\end{figure*}

Figure \ref{fig:Approach} shows our approach to solve the propeller design optimization problem. It is a hybrid approach when ML model is fused with traditional algorithm with numerical physics in loop of optimization. 

During training time, we train our random forest and decision tree. For training both models, we used requirement data ($r \sim R$) as an input and the corresponding geometric design values($g \sim G$) and resulting efficiency ($\eta$) forming a tuple as output.  The random forest is trained to learn the inverse regression and predict the design geometry along with efficiency on a given requirement. The decision tree on the other hand does inverse mapping from requirement space to design geometry and efficiency searched during data generation.  The goal of random forest is to learn a function $ f : \mathcal{R} \mapsto \mathcal{G},\eta $ that is continuous so that we can regress for in between points however, the decision tree is memory map and just does space partitioning on seen data. Using both gives up good quality seed initial design. Since we do not know possible efficiency that can be achieved on given requirement, we possible take all possible prediction and sort on bases on efficiency to get the best design found yet. 
Direct prediction of random forest is an average of all geometric design and efficiency corresponding to the given requirement, which may or may not be a very good initial design. Here the role of random forest is generalisation and regression on unseen data. The role of decision tree is to does non-linear one to many inverse mapping. We selected all the designs that are on the leaf of decision tree and include those as well to our baseline designs- this is called baseline prediction. Using both models we get good quality initial seed designs.
In the next stage, we take these baseline designs as initial population and start the genetic algorithm search for the final optimized design.  (refer to fig \ref{fig:Approach}). 

In GA, chromosomes are represented by arrays of bits or character strings that have an optimization function encoded in them. Strings are then processed by genetic operators, and the fittest candidates are chosen. We run GA in loop with openProp numerical simulator until budget.

\subsection{Data generation \& Training}
For data generation, we took the design space used by \cite{vardhan2021machine}. The geometric design space is of the order of $10^{27}$ (diameter * nine alternative chord radial profiles), whereas the requirement space after coarse discretization is on the order of $10^{11}$ (thrust x velShip x RPM) with combined search space is $10^{38}$.
We take a single sample point from the physical design space and the requirement space and input it into the OpenProp optimizer. OpenProp internally optimizes this design using iterative numerical methods and computes the performance metric ($\eta$). We used this 0.205 million valid design data point for our training and testing. Using this design corpus, we trained both random forest regression\cite{breiman2001random} model and the decision tree. Other hyperparameters of the random forest model are an ensemble of $100$ decision trees with mean squared error as splitting criteria of the node. For the decision tree model, we chose squared error as the splitting criteria of the node, and nodes are expanded until all leaves are pure. Other hyperparameters are kept as default settings as in SKlearn \cite{pedregosa2011scikit}.

\section{Experiment and Results}
\label{sec:results}
For sharing the result, we have two things to share: 
\begin{enumerate}
    \item prediction accuracy of random forest on test data. 
    \item Empirical evaluation of SAO (on example design optimization problems and its comparison with baseline (Genetic Algorithm). 
\end{enumerate}
For testing the prediction accuracy of our trained model we selected 5\% of data randomly from the dataset. To assess the quality of prediction, we used the following common statistics as evaluation metrics: 
\begin{enumerate}
    \item average \textbf{residual}, $\Delta Z= (\eta_{truth}- \eta_{predicted})/\eta_{truth}$ per sample 
    \item the \textbf{accuracy}, percentage of the number of samples whose residual is within acceptable error of 5\% i.e $|\Delta Z| <0.05$. 
\end{enumerate}
It measures the percentage of test data on which the prediction of efficiency is within 5\% of error  (since efficiency is a good metric and target of final prediction). 
We found percentage prediction accuracy on test data for the random forest is around 90\%. For the decision tree, we fitted it with the entire  data, since we just want space partitioning of collected data. 

\begin{figure*}[h!]
    \centering
    \begin{tabular}{|c|c|}
        \hline
        \hypertarget{fig:optsamples:BOEI}{}
        \hypertarget{fig:optsamples:BOLCB}{}
        
        & \\
          \includegraphics[width=0.45\textwidth]{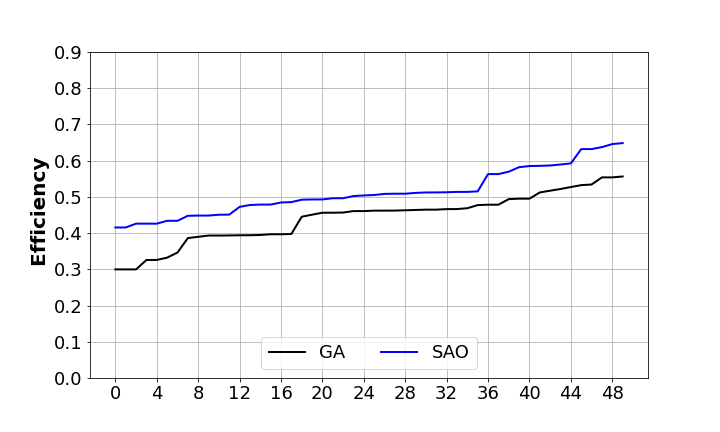} & 
          \includegraphics[width=0.45\textwidth]{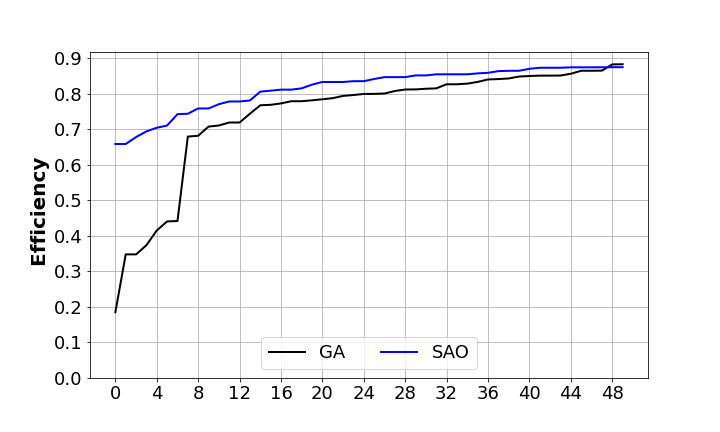}\\
          (a) &  (b)   \\
          \hline
            \hypertarget{fig:optsamples:GA}{}
            \hypertarget{fig:optsamples:NM}{}
          & \\
            \includegraphics[width=0.45\textwidth]{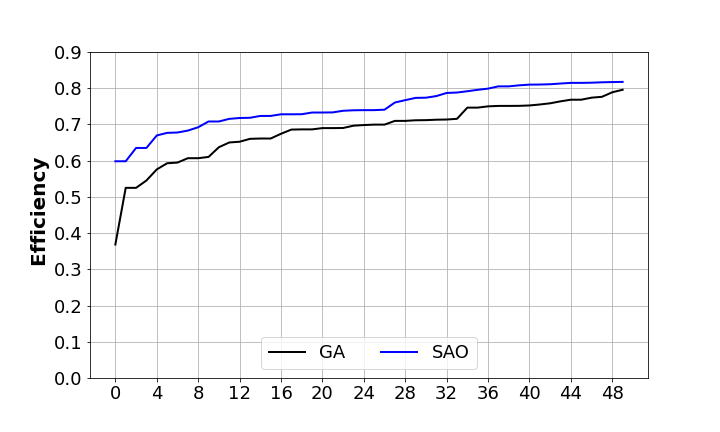} &
            \includegraphics[width=0.45\textwidth]{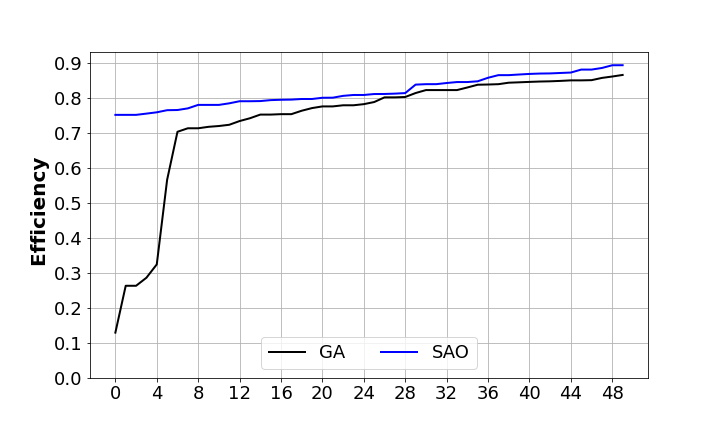} 
            \\
          (c) &  (d) \\
        \hline
        \hypertarget{fig:optsamples:maximin}{}
        \hypertarget{fig:optsamples:VMC}{}
        & \\
         \includegraphics[width=0.45\textwidth]{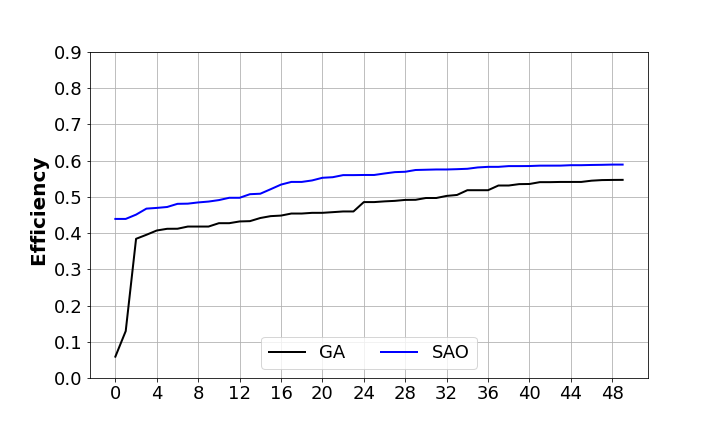}
        & \includegraphics[width=0.45\textwidth]{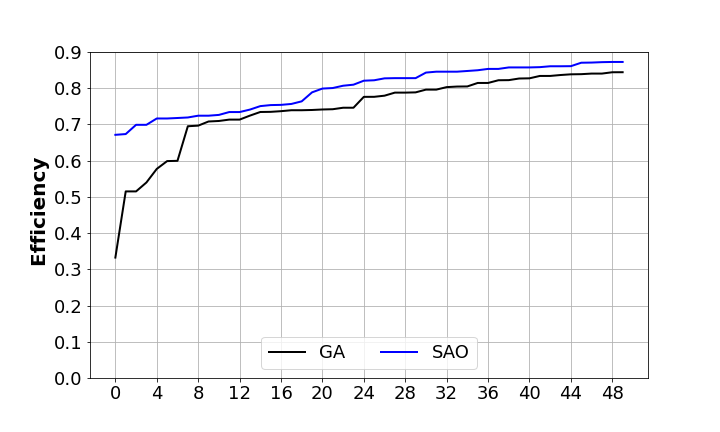} \\
        (e) &  (f) \\
         \hline
        \hypertarget{fig:optsamples:gg}{}
        \hypertarget{fig:optsamples:hh}{}
        & \\
         \includegraphics[width=0.45\textwidth]{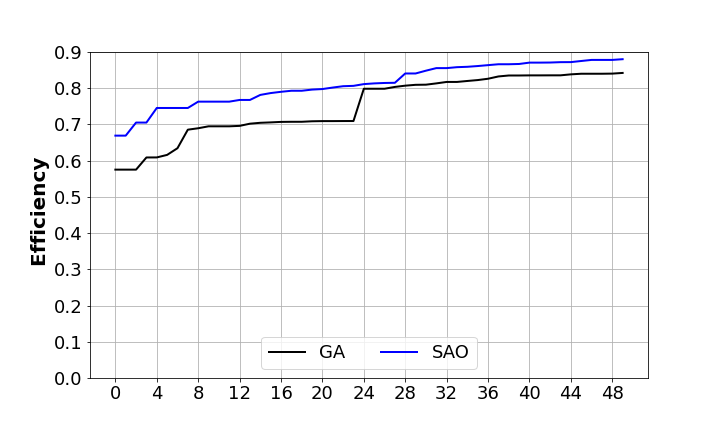}
        & \includegraphics[width=0.45\textwidth]{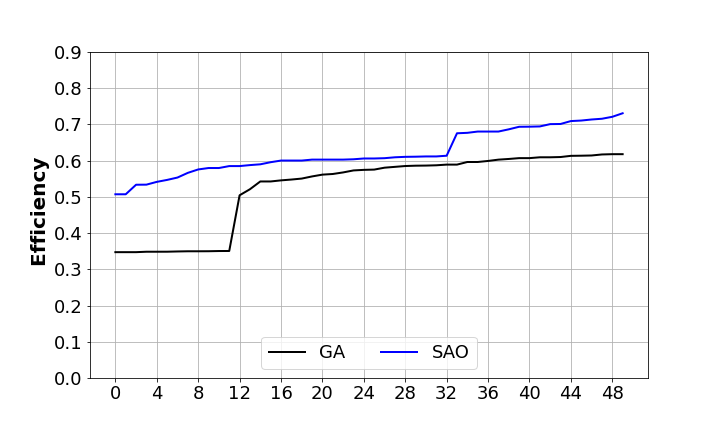} \\
        (g) &  (h) \\
\hline
    \end{tabular}
    \caption{Results of sample optimization runs using GA and Surrogate Assisted Optimization (SAO): Due to learned manifold SAO provided better seed design for evolutionary optimization and get better performing design in given budget. Requirements for optimization are sampled randomly for Design space:\{thrust (Newton) ,velocity of ship (m/s), RPM\}\protect\hyperlink{fig:optsamples:BOEI}{(a)} \{51783,7.5,3551\}, \protect\hyperlink{fig:optsamples:BOLCB}{(b)} \{127769,12.5,699\}, \protect\hyperlink{fig:optsamples:GA}{(c)} \{391825,12.5,719\}, \protect\hyperlink{fig:optsamples:NM}{(d)} \{205328,19.5,1096\}, \protect\hyperlink{fig:optsamples:maximin}{(e)} \{301149,7.5,1215\}, \protect\hyperlink{fig:optsamples:VMC}{(f)}  \{314350,16.0,777\}, \protect\hyperlink{fig:optsamples:gg}{(g)}  \{31669,17.5,2789\}, \protect\hyperlink{fig:optsamples:hh}{(h}  \{476713,15.5,2975\}.}
    \label{fig:emp_eval}
\end{figure*}

For the empirical evaluation of SAO, we chose Genetic Algorithm as our baseline optimization algorithm that is frequently deployed in such situations. Figure \ref{fig:emp_eval} shows the evaluation traces of the optimization process. It can be observed that due to the trained surrogate, we get a better initial seed design, and further optimization in the second step using GA provides better designs on the given budget in comparison to applying GA which starts with a random seed design.   
\section{Related Works}
\label{sec:related}
ML has the ability to learn from raw data  and its wide application in design and operation is shown in various works \cite{liang2018deep,madani2019bridging,vardhan2021rare,abbeel2010autonomous,vardhan2023search}. The optimization in the design process is also changing from traditional model-based optimization \cite{vardhan2019modeling} to the availability of ML-based cheap surrogate that can replace the traditional first principle physics-based models\cite{koziel2013surrogate} or by directly solving inverse problems\cite{tarantola2005inverse,tarantola2006popper,aller2023study}.
Lee et al \cite{lee2004optimized, calcagni2010automated} used a genetic algorithm for optimizing the propeller design.  However, the application of AI and ML in real-world system design is relatively slow. \cite{vardhan2021machine,doijode2022application} are a few known works to apply AI-ML concepts in the design of propellers. 
\section{Conclusion and Future Work}
\label{sec:conclusion}
We showed that even in high-dimensional design optimization problems, SAO can speed up the design optimization process. By adding more data, it would be possible to improve further. The future work in this direction would be adding more data to ML models and seeing what is the maximum performance that can be achieved. Based on our intuition we hope that it is possible to find an optimal design in $O(1)$ time complexity if a sufficient amount of data is collected and models are trained on it.  

%
%
 \bibliographystyle{splncs04}
\bibliography{References}

\end{document}